\documentclass{article}
\usepackage{iclr2027_conference,times}
\iclrfinalcopy
\usepackage[letterpaper,left=1.5in,right=1.5in,top=1in,bottom=1in]{geometry}
\renewcommand{\headrulewidth}{0pt}
\usepackage[T1]{fontenc}
\usepackage[utf8]{inputenc}
\usepackage{amsmath,amssymb,amsthm,booktabs,array,tabularx}
\newtheorem{proposition}{Proposition}
\DeclareMathSizes{8.3}{8.3}{6}{5}
\usepackage{graphicx,xcolor}
\usepackage{placeins,microtype}
\usepackage{hyperref,url}
\hypersetup{colorlinks=true,linkcolor=blue!40!black,citecolor=blue!40!black,urlcolor=blue!40!black,pdftitle={Preview-Guided Memory: Learning to Reread for Long-Context Reasoning},pdfauthor={Xirui Liang, Luotong Wang, Tianchen Zhao, Tengxuan Liu, Chao Xu, Hanting Chen, Yu Wang}}

\title{Preview-Guided Memory: Learning to Reread for Long-Context Reasoning}
\author{%
Xirui Liang$^{1}$ \quad Luotong Wang$^{2}$ \quad Tianchen Zhao$^{2}$ \quad Tengxuan Liu$^{2}$ \\
\bfseries Chao Xu$^{1}$ \quad Hanting Chen$^{1}$\thanks{Corresponding author.} \quad Yu Wang$^{2}$ \\[4pt]
\normalfont $^{1}$Peking University \quad $^{2}$Tsinghua University
}
\begin{document}
\maketitle
\fancyhead{}
\begin{abstract}
LLM agents accumulate long contexts as they read documents, use tools, and carry out multi-step tasks. Memory-based methods handle these contexts by reading text in chunks and maintaining a compact memory. However, single-pass readers must decide what to retain before seeing later chunks, while still processing every chunk regardless of its relevance. We propose Preview-Guided Memory (PGM), which uses a preliminary preview to guide both reading and memory updates. In the first pass, PGM generates a short, question-independent preview of each chunk and groups neighboring previews into pages. A learned selector uses these pages and the question to choose which chunks to read in detail in the second pass. The reader then processes the selected chunks in order, using the corresponding preview page when updating its memory. These previews provide clues about upcoming content, while selection reduces the number of memory updates. We keep the reader fixed and train the selector with supporting-fact supervision followed by reinforcement learning, rewarding answer quality and evidence coverage, with a bonus for reduced reading when the answer is correct. PGM-Selective-7B achieved state-of-the-art on both HQA and longbench v2 with fewer tokens for reasoning.
\end{abstract}

\section{Introduction}
\label{sec:intro}

LLM agents often work with long contexts containing documents, tool outputs, and previous interactions~\citep{yao2023react,packer2023memgpt}. As a task continues, this context grows, and information collected early may still be needed much later. Keeping everything in the prompt is straightforward, but repeatedly processing it becomes expensive and eventually reaches the model's context limit. Larger context windows allow longer inputs~\citep{ding2024longrope,yang2025qwen1m}, but do not remove the cost of reading them. Long-context reasoning therefore calls for methods that can retain useful information without keeping all previous text in the active context.

Memory-based methods address this problem by compressing previously read text into a short memory. MemAgent~\citep{yu2026memagent}, for example, divides a document into chunks and reads them sequentially. At each step, the model receives the question, the current chunk, and the previous memory, then writes an updated memory for the next step. After reading the document, it answers from the final memory. ReMemR1~\citep{shi2026rememr1} allows the reader to revisit historical memories during reading. These methods keep individual model calls within a limited context window while carrying information across much longer documents.

However, reading and compressing one chunk at a time has two limitations. First, the model must decide what to keep before it has seen the content that comes next. Although the question tells it what to look for, the role of a detail may only become clear after reading another passage. A name or relation that appears unimportant in the current chunk may later be needed to connect two pieces of evidence. If that detail was never stored, revisiting historical memories may not recover it. Second, reading every chunk in detail requires repeated memory-update calls, even when many chunks contribute little to the answer. Each call still includes the question and accumulated memory. The reader therefore makes compression decisions with incomplete information about what lies ahead, while spending repeated computation on content it may not need.

We propose Preview-Guided Memory (PGM), which first skims a document to produce short, question-independent previews and then uses them to guide a second pass of reading. The previews serve two purposes. They give the reader clues about surrounding content, including passages not yet read in detail, when deciding what to retain in memory. They also help a learned selector use the question to choose which chunks to read. The reader then processes the selected chunks in their original order, using the previews to guide its memory updates. Preview generation still covers the full document; selection reduces the subsequent reading and memory updates. We keep the reader fixed and train the selector with supporting-fact supervision followed by reinforcement learning from the reader's final answers.

We evaluate PGM on HQA and LongBench v2. PGM-Selective-7B achieves higher accuracy with fewer tokens across both benchmarks. These results demonstrate that preview-guided selective reading improves the effectiveness and efficiency of memory-based reasoning across long contexts and diverse tasks.

\section{Related Work}
\label{sec:related}

\paragraph{Memory-based long-context reasoning.}
Memory-based agents organize past information through memory stores or compact states~\citep{packer2023memgpt,zhong2023memorybank,chhikara2025mem0,zhou2025mem1}. For document reasoning, MemAgent incrementally updates a textual memory~\citep{yu2026memagent}; ReMemR1 revisits historical memories~\citep{shi2026rememr1}; and InfMem and MemReread retrieve or reread source evidence~\citep{wang2026infmem,ji2026memreread}. PGM gives memory updates a view of neighboring content before it is read in detail, helping the reader anticipate which information to retain.

\paragraph{Retrieval and multi-pass reading.}
Retrieval-augmented methods select evidence to support generation~\citep{karpukhin2020dpr,lewis2020rag,izacard2021fid,borgeaud2021retro,shi2024replug}. Self-RAG learns retrieval and reflection, while RAPTOR retrieves over hierarchical summaries~\citep{asai2023selfrag,sarthi2024raptor}. Closer to PGM, DOS RAG preserves source order~\citep{laitenberger2025dosrag}, ReadAgent uses gists for raw-page lookup~\citep{lee2024readagent}, and ReadTwice shares first-pass representations across segments~\citep{zemlyanskiy2021readtwice}. PGM reuses the same textual previews to select raw chunks and guide subsequent memory updates.

\paragraph{Learning reading decisions.}
RL has been used to optimize long-context reasoning and search policies~\citep{wan2025qwenlongl1,jin2025searchr1}. LycheeMemory selects compressed key--value blocks~\citep{chen2026lycheememory}, while GRU-Mem learns when to update memory or stop reading~\citep{sheng2026grumem}. PGM initializes its selector with supporting facts, then uses the final answers of a fixed reader to evaluate complete chunk selections. LycheeMemory's supervised gate uses the question and evolving memory; PGM reuses textual previews for both selection and memory updates.

\section{Method}
\label{sec:method}
\label{sec:formulation}

\paragraph{Sequential memory as a starting point.}
MemAgent~\citep{yu2026memagent} handles a long document by compressing previously read text into a bounded memory. Given a question $q$ and ordered source chunks $D=(c_1,\ldots,c_N)$, its reading process can be written as
\begin{equation}
 \begin{aligned}
 m_i^{\mathrm{seq}}&=U_{\mathrm{seq}}(q,c_i,m_{i-1}^{\mathrm{seq}}),
 \quad i=1,\ldots,N,\\
 \hat y_{\mathrm{seq}}&=F_{\mathrm{seq}}(q,m_N^{\mathrm{seq}}),
 \qquad |m_i^{\mathrm{seq}}|\le B.
 \end{aligned}
 \label{eq:sequential-reader}
\end{equation}
Starting from empty memory, each update rewrites the accumulated information within a $B$-token budget, and the final answer uses only the question and the last memory. This keeps individual calls short. However, a fact needed later must survive every intervening rewrite, even before the reader knows how it connects to the question.

\begin{figure}[t]
\centering
\includegraphics[width=\linewidth]{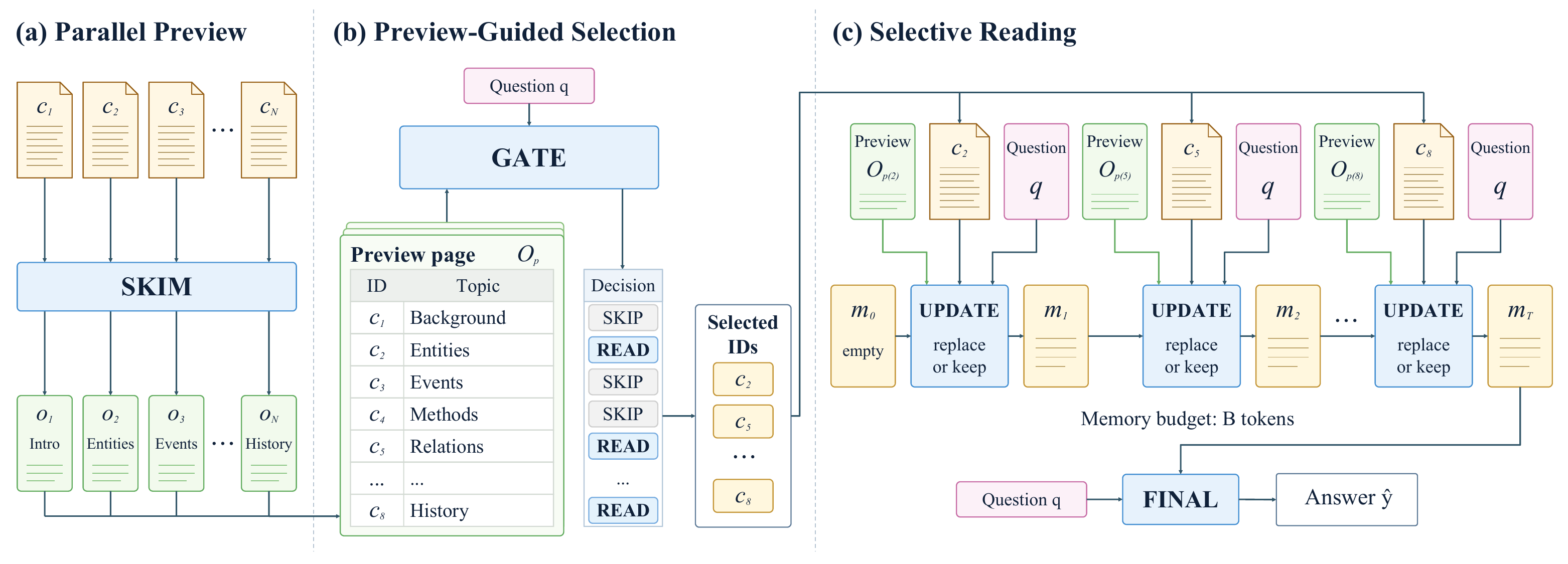}
\caption{\textbf{Overview of Preview-Guided Memory (PGM).}
(a) In the first pass, SKIM generates indexed, question-independent topical previews for all chunks in parallel. Neighboring previews are grouped into pages.
(b) GATE uses the question and preview pages to select raw chunks before the second pass.
(c) Starting from empty memory $m_0$, the reader processes the selected chunks in source order and updates memory within a $B$-token budget, using the corresponding preview page as context.
The same previews guide both chunk selection and memory updates.
FINAL produces $\hat y$ using only the question $q$ and final memory $m_T$.
Topic labels and selections are schematic; green arrows denote preview conditioning.}
\label{fig:overview}
\end{figure}

\paragraph{Preview, then read and remember.}
These limitations suggest that finding evidence and retaining it should be considered together. PGM first builds short, question-independent previews, giving reading decisions a view of the document before detailed memory updates begin. The same previews then guide \emph{which chunks to read} and \emph{what to retain} during those reads. Selection shortens the sequence of recurrent updates, while neighboring previews provide context for interpreting the current passage. Figure~\ref{fig:overview} presents the general workflow.

\subsection{Preview before detailed reading}
\label{sec:previews}

Let $\theta$ and $\phi$ denote the reader and selector parameters. Before updating memory, PGM builds a compact view of the source:
\begin{equation}
 o_i=H_\theta(c_i),\qquad
 O_p=\{(i,o_i):i\in I_p\},
 \label{eq:preview}
\end{equation}
where $o_i$ is a topical preview and $I_p$ contains the indices of neighboring chunks grouped into page $p$. We write $p(i)$ for the page containing chunk $i$ and $O$ for all preview pages. Each page fits within a bounded input budget, so the preview representation can grow with the document without requiring a single long-context call.

Each preview describes a chunk's topics and points back to the raw source. Generating it without the question separates document previewing from task-specific selection. Neighboring previews reveal how topics develop across chunks; each preview depends only on its own chunk, so all first-pass calls can run in parallel. These reusable previews are lossy and may omit answer-bearing details; contextual cues guide retention, while raw chunks supply the detailed evidence.

\subsection{Use the preview to select and remember}
\label{sec:preview-use}

\paragraph{Select where to read.}
\label{sec:gate}
Once the previews are available, the question determines which parts warrant detailed reading. The selector receives one preview page at a time and scores its entries together in a single forward pass:
\begin{equation}
 a_i=\sigma\!\left(\ell_\phi(q,O_{p(i)},i)\right),
 \label{eq:gate}
\end{equation}
where $\ell_\phi$ is the selection logit and $\sigma$ is the sigmoid function. These scores determine the selected set $S=\{s_1,\ldots,s_T\}$, with $s_1<\cdots<s_T$. All page decisions are completed before the second pass. The selector therefore chooses the reading sequence from the question and previews, while the reader subsequently processes that sequence in source order.

\paragraph{Decide what to remember.}
\label{sec:update}
Selection is only the first use of the previews. A relevant chunk may contain several details, and the reader still has to decide which ones deserve space in memory. PGM supplies the corresponding preview page again at each update:
\begin{equation}
 \begin{aligned}
 m_t &= U_\theta(q,c_{s_t},m_{t-1},O_{p(s_t)}),\\
 \hat y &= F_\theta(q,m_T), \qquad |m_t|\leq B.
 \end{aligned}
 \label{eq:reader}
\end{equation}
Starting from empty memory $m_0$, the reader either replaces its memory with an updated version or keeps it unchanged. This single memory carries across preview pages; after the last selected chunk, the final answer is generated from the question and $m_T$.

The inputs have complementary roles: previews provide surrounding topic cues, raw chunks supply new evidence, and memory carries earlier facts across selected chunks and page boundaries. The update prompt treats previews as navigation rather than evidence: old facts must come from the previous memory, and new facts from the current raw chunk. A preview can therefore guide retention without itself supplying the answer or a missing relation.

We call this workflow PGM-Selective. PGM-ReadAll retains the same previews and memory-update procedure but reads every source chunk. This full-reading reference lets us examine the benefit of selection within the preview-guided workflow.

\begin{figure}[t]
\centering
\includegraphics[width=\linewidth]{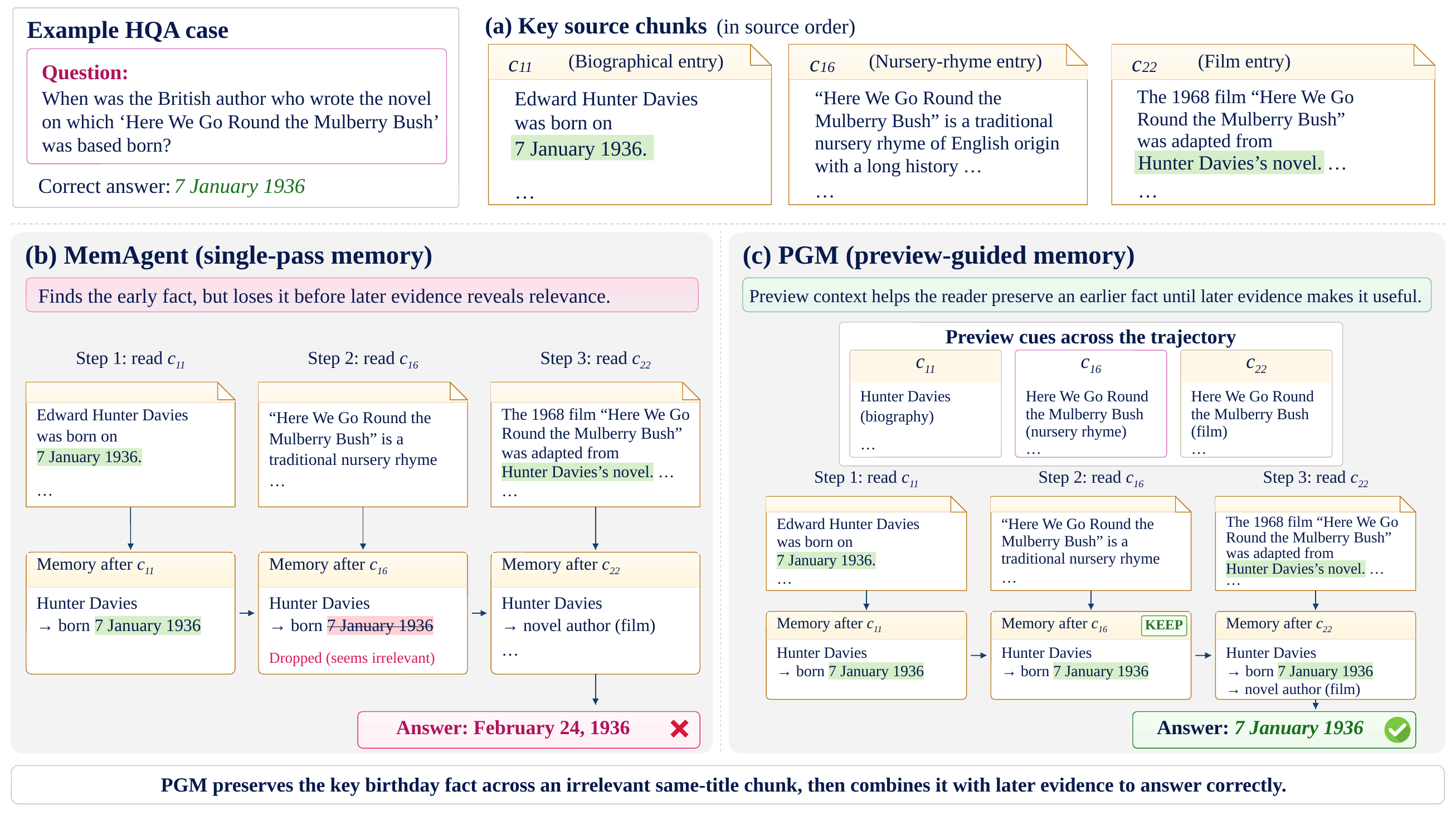}
\caption{\textbf{Motivating HQA case.}
(a) Key source chunks. (b) MemAgent loses an initially stored birth date before the film entry identifies the author. (c) PGM-Selective retains the date through the same-title distractor and combines it with the later author relation.
Preview cues are summarized across pages, not reproduced from one page; the film preview supplies neither the author relation nor the date.
Only key states are shown; intermediate updates are omitted. PGM reads seven of 23 chunks in the second pass.
Appendix~\ref{app:case-study} gives the run settings and local rewrite control for this illustrative diagnostic pair.}
\label{fig:hqa-case}
\end{figure}

\paragraph{When relevance becomes clear too late.}
Figure~\ref{fig:hqa-case} compares PGM with MemAgent on a real 112K-token HQA trajectory. The question asks for the birth date of the author whose novel was adapted into \emph{Here We Go Round the Mulberry Bush}. MemAgent stores Hunter Davies's birth date, \emph{7 January 1936}, at chunk~11, but drops it while processing a same-title nursery-rhyme entry at chunk~16. When chunk~22 identifies Davies as the novel's author, the date is absent and its final answer is incorrect. The fact was extracted, then lost before later evidence established its relevance.

On the same question and source document, PGM retains Davies's birth date, combines it with the later author relation, and answers correctly after reading seven of the 23 chunks in its second pass. Between chunks~11 and~22, MemAgent performs ten intervening updates, whereas PGM reads only chunks~13 and~16 before reaching chunk~22. PGM therefore succeeds while still processing the same-title distractor, rather than by excluding every distractor. Its preview page at chunk~16 lists the later film entry but not the author relation or birth date; those facts come from the raw source. Appendix~\ref{app:case-study} preserves the run settings and local rewrite control.

\subsection{Training the reader and selector}
\label{sec:training}

\paragraph{From full reading to selective reading.}
We first train PGM-ReadAll with SFT followed by RL, then initialize selective training from this checkpoint. Selective SFT learns chunk selection from supporting facts, followed by Selective RL using final-answer feedback. Thus, the Selective SFT-only ablation already inherits both ReadAll training stages.

\paragraph{Start by locating evidence.}
We align supporting facts with the source chunks and assign $y_i^{\mathrm{ev}}=1$ to chunks that contain them. A weighted binary cross-entropy objective initializes the selector:
\begin{equation}
 \mathcal L_{\mathrm{ev}}=-\frac1N\sum_{i=1}^N
 \left[w_+y_i^{\mathrm{ev}}\log a_i+
 (1-y_i^{\mathrm{ev}})\log(1-a_i)\right].
 \label{eq:bce}
\end{equation}
The positive-class weight $w_+>1$ emphasizes retaining evidence. This provides a direct learning signal before the selector has learned to compare complete reading trajectories.

\paragraph{Evaluate a set through its final answer.}
For each training question, we sample $K>1$ chunk sets. Each set produces its own source-ordered memory sequence and final answer through Equation~\ref{eq:reader}. Writing $z_i=\mathbf1[i\in S]$, the log probability of a sampled selection is
\begin{equation}
 \log\pi_\phi(S\mid q,O)=\sum_{i\in\mathcal I}
 \left[z_i\log p_i+(1-z_i)\log(1-p_i)\right],
 \label{eq:set-probability}
\end{equation}
where $\mathcal I$ contains the stochastic selection decisions and $p_i$ is a smoothed version of $a_i$. This probability includes both reading and skipping decisions across preview pages. We optimize the expected reward $\mathcal J(\phi)=\mathbb E_{S\sim\pi_\phi}[R(S)]$, so feedback is attached to the complete selection rather than to isolated relevance scores.

The reward combines answer quality, evidence coverage, and economical reading. Let $C$ be normalized answer exact-match correctness, $F$ answer F1, and $C_{\mathrm{all}}$ the correctness of the paired PGM-ReadAll answer. For each of the $J$ supporting facts, let $g_j(S)$ indicate whether the selected chunks contain at least one complete occurrence. We measure the fraction of facts covered, whether all facts are covered, and the fraction of source tokens skipped as
\begin{equation}
 G(S)=\frac1J\sum_{j=1}^Jg_j(S),\qquad
 E(S)=\prod_{j=1}^Jg_j(S),\qquad
 s(S)=1-\frac{\sum_{i\in S}|c_i|}{\sum_{i=1}^N|c_i|}.
 \label{eq:coverage}
\end{equation}
The trajectory reward is
\begin{equation}
 \begin{aligned}
 R(S)={}&\alpha C-\beta C_{\mathrm{all}}(1-C)+\gamma(1-C)F+\delta G(S)\\
        &-\eta\bigl(1-E(S)\bigr)+\lambda_u C\,s(S)-\zeta V,
 \end{aligned}
 \label{eq:reward}
\end{equation}
where $V$ indicates an invalid answer format. The nonnegative coefficients weight correctness, regression from full reading, partial-answer credit, coverage, missing evidence, and invalid formatting, respectively; $\lambda_u$ controls the reading bonus at training update $u$.

Each part addresses a different consequence of selection. Answer feedback measures whether the evidence remains usable after memory updates, with an additional penalty for losing an answer that full reading gets right. Coverage rewards retaining the links needed for multi-hop reasoning. Crucially, the reading bonus is multiplied by $C$: shorter reading is rewarded when it preserves a correct answer. The selector thus learns to remove unnecessary reading while retaining evidence that the downstream reader can use.

\paragraph{Compare alternative reading sequences.}
For rewards $R_1,\ldots,R_K$ from the same question, a leave-one-out baseline compares each selection with its alternatives~\citep{williams1992reinforce,ahmadian2024rloo}:
\begin{equation}
 A_k=R_k-\frac{1}{K-1}\sum_{\ell\ne k}R_\ell.
 \label{eq:loo}
\end{equation}
A selection that outperforms the other samples receives a positive advantage; a weaker selection receives a negative one. We use these advantages as fixed policy-gradient weights in the regularized objective
\begin{equation}
 \mathcal L=-\frac1K\sum_{k=1}^K A_k\log\pi_\phi(S^{(k)}\mid q,O)
 +\lambda_{\mathrm{KL}}\overline D_{\mathrm{KL}}+\lambda_{\mathrm{ev}}\mathcal L_{\mathrm{ev}}
 -\lambda_{\mathrm{ent}}\overline{\mathcal H}.
 \label{eq:loss}
\end{equation}
Here, $\overline D_{\mathrm{KL}}$ anchors the policy to the supervised selector and $\overline{\mathcal H}$ encourages exploration; both are averaged over chunks, with nonnegative regularization weights. Retaining $\mathcal L_{\mathrm{ev}}$ preserves the direct evidence signal while answer feedback refines set-level decisions. The trained selector changes which chunks enter the memory-update sequence, allowing the same reader to operate on a more useful selection. Appendix~\ref{app:training} gives the label alignment and training settings.

\subsection{Why previewing before compression can help}
\label{sec:theory}

The distinction between Equations~\ref{eq:sequential-reader} and~\ref{eq:reader} is the timing of information: a preview may reveal which earlier detail will become relevant before that detail is compressed. We formalize this through a delayed-reference task, using a classical random-access coding argument~\citep{nayak1999randomaccess}. A $B$-token memory over a finite vocabulary has a finite state space; $b$ below measures its capacity in bits.

\begin{proposition}[A memory advantage from an early dependency cue]
\label{prop:preview-memory}
Let the first of two chunks contain $d$ independent uniform bits $X=(X_1,\ldots,X_d)$, and let the second contain an independent uniform index $Z\in\{1,\ldots,d\}$. The fixed question asks for $Y=X_Z$. A single-pass reader retains a state $W$ with at most $2^b$ possible values after the first chunk and has no other access to $X$. Its minimum average answer error $e$ satisfies
\begin{equation}
 b\ge d\bigl[1-h_2(e)\bigr],
 \label{eq:memory-lower-bound}
\end{equation}
where $h_2(e)=-e\log_2e-(1-e)\log_2(1-e)$ is binary entropy. In contrast, if the preview page exposes $Z$ before the first memory update, a preview-guided reader can answer exactly using only two memory states.
\end{proposition}

\begin{proof}
Let $e_j$ be the optimal error for recovering $X_j$ from $W$. Because $Z$ is uniform and independent of $(X,W)$, $e=d^{-1}\sum_j e_j$. Binary decoding gives $H(X_j\mid W)\le h_2(e_j)$: given $W$, the predicted bit and its error indicator determine $X_j$. Independence of the source bits, conditional entropy subadditivity, and concavity of $h_2$ imply
\begin{equation}
 \begin{aligned}
 b\ge H(W)\ge I(X;W)
 &=d-H(X\mid W)\\
 &\ge d-\sum_{j=1}^d h_2(e_j)
 \ge d\bigl[1-h_2(e)\bigr].
 \end{aligned}
 \label{eq:memory-bound-proof}
\end{equation}
For the preview-guided construction, group the two previews on one page and let the second preview report $o_2=Z$. When reading the first raw chunk, write $X_Z$ into memory; keep that bit during the second update and output it at the end. The first preview need not retain any of the bits in $X$.
\end{proof}

With fewer than $d$ bits of retained information, the single-pass reader cannot attain zero error on this task. The preview-guided construction retains only the needed bit: the preview identifies \emph{which} detail matters, while its value still comes from the raw chunk. Both readers can process the same two chunks, so the separation isolates preview-conditioned memory writing from chunk selection. In PGM, a neighboring preview can play this role by exposing an entity or relation needed later. If that dependency cue is decoded incorrectly with probability at most $\varepsilon$, the same construction has error at most $\varepsilon$.

\paragraph{Accuracy takes precedence in the selection reward.}
The reading bonus also has an exact property that follows from the objective, independently of how the selector is optimized.
\begin{proposition}[Correctness-prioritized trajectory reward]
\label{prop:reward-priority}
Suppose all reward coefficients are nonnegative and
$\alpha>\gamma+\delta+\eta+\zeta$.
For any two trajectories of the same question, with $C(S^+)=1$ and $C(S^-)=0$, Equation~\ref{eq:reward} satisfies
\begin{equation}
 R(S^+)-R(S^-)\ge
 \alpha-\eta-\zeta-\gamma-\delta>0.
 \label{eq:reward-priority}
\end{equation}
This ordering holds for every nonnegative reading-bonus weight $\lambda_u$.
\end{proposition}
\begin{proof}
For a correct answer, the regression and partial-answer terms vanish, and $R(S^+)\ge\alpha-\eta-\zeta$. For an incorrect answer, the reading bonus vanishes, the remaining penalties are nonpositive, and $R(S^-)\le\gamma+\delta$. Subtracting proves the claim.
\end{proof}
The training coefficients in Appendix~\ref{app:training} satisfy this condition. Thus, a shorter failed trajectory cannot outrank a successful one through the reading bonus; among otherwise equivalent successful trajectories, an active bonus favors less reading.

\subsection{How previewing changes the reading cost}
\label{sec:token-analysis}

Previewing adds an initial pass, but allows the second pass to contain fewer recurrent updates. To see this trade-off, consider MemAgent, which updates memory on all $N$ source chunks~\citep{yu2026memagent}. Let $c_m$ be its representative input-plus-output token cost per update. For PGM, let $c_s$, $c_g$, and $c_u$ be the corresponding costs of generating one preview, scoring one preview page, and updating memory on one selected chunk. With $M$ preview pages and $T=|S|$ selected chunks,
\begin{equation}
 \begin{aligned}
 C_{\mathrm{MemAgent}} &\simeq N c_m+C_f^{\mathrm{MemAgent}},\\
 C_{\mathrm{PGM}} &\simeq N c_s+M c_g+T c_u+C_f^{\mathrm{PGM}},
 \end{aligned}
 \label{eq:cost-comparison}
\end{equation}
where $C_f$ denotes final-answer cost. Each term includes its prompt and output tokens. Under fixed chunk, preview-page, memory, and generation budgets, both workflows scale as $O(N)$.

Writing $\rho=T/N$, and neglecting the final-answer terms for long documents, gives
\begin{equation}
 \frac{C_{\mathrm{PGM}}}{C_{\mathrm{MemAgent}}}
 \simeq\frac{c_s}{c_m}+\frac{M}{N}\frac{c_g}{c_m}
 +\rho\frac{c_u}{c_m}.
 \label{eq:relative-cost}
\end{equation}
The first two terms build and use the preview index; the last pays for detailed reading. PGM previews every chunk but concentrates recurrent updates on the selected sequence. The experiments measure the full cost.

The comparison with PGM-ReadAll further isolates the savings from selection. In the same representative-cost model, full reading uses $N c_s+N c_u+C_f^{\mathrm{ReadAll}}$ tokens, so
\begin{equation}
 C_{\mathrm{ReadAll}}-C_{\mathrm{PGM}}
 \simeq (N-T)c_u-Mc_g+C_f^{\mathrm{ReadAll}}-C_f^{\mathrm{PGM}}.
 \label{eq:readall-saving}
\end{equation}
The shared preview pass cancels. With similar final-answer costs, selection saves tokens when the skipped updates outweigh the selector's $Mc_g$ overhead. Savings thus come from reducing recurrent reading and rewriting after the initial scan.
\section{Experiments}
\label{sec:experiments}

We evaluate PGM on HQA and LongBench v2 to study answer quality and end-to-end token use. We compare against long-context and memory-based baselines, analyze inference cost, and examine the effects of training and chunk selection.

\subsection{Experimental setup}
\label{sec:setup}

\paragraph{Benchmarks.}
We evaluate HQA at 28K, 112K, 224K, 448K, and 896K tokens, with 128 questions at each length~\citep{yang2018hotpotqa,yu2026memagent}. We report substring exact match (SubEM) and its equal-weight average over the five lengths. LongBench v2 contains 503 multiple-choice questions across six task categories~\citep{bai2025longbenchv2}. We report overall accuracy and breakdowns by task category, context length, and difficulty.

\paragraph{Baselines.}
We compare with Qwen2.5 and Qwen2.5-1M~\citep{yang2025qwen25,yang2025qwen1m}, BM25 and dense DOS RAG~\citep{laitenberger2025dosrag}, and the memory-based methods ReadAgent, MemAgent, LycheeMemory, ReMemR1, InfMem, and MemReread, discussed in Section~\ref{sec:related}. PGM-ReadAll uses the same previews and memory-update procedure as PGM-Selective, but reads every chunk in the second pass. We also compare BM25 and dense retrieval with the frozen PGM reader under matched reading budgets. Raw-text selection uses individual chunks or five-chunk windows with a retrained preview-free reader; this reader also defines the no-preview ReadAll baseline. Implementation details are provided in Appendix~\ref{app:implementation}.

\paragraph{Token cost.}
We count all input and output tokens for preview generation, selection, memory updates, and final answering. HQA cost is averaged over the same five lengths as SubEM and normalized to Qwen2.5-7B-1M. In the ablations, the reading ratio is the fraction of raw-source tokens read in the second pass.

\subsection{Main results}
\label{sec:answer-quality}

\paragraph{HQA.}
Table~\ref{tab:long} shows that PGM-Selective-7B achieves 78.75 average SubEM, compared with 75.00 for PGM-ReadAll-7B, 76.72 for ReMemR1, and 75.78 for MemAgent. It matches or exceeds the strongest non-PGM baseline at every evaluated length. The gains over ReadAll are largest at 448K and 896K, reaching 10.16 and 6.25 percentage points. Selection thus improves answer quality while reducing the number of chunks passed through the memory-update sequence.

Across the 32-fold increase from 28K to 896K, Selective declines from 82.81 to 75.00 SubEM, compared with 82.03 to 68.75 for ReadAll. The advantage at the two longest lengths is consistent with prioritizing a useful reading sequence over repeatedly processing every chunk.

With the same PGM reader, BM25 and dense retrieval average 72.66 and 57.66 SubEM. Raw single-chunk, neighboring-chunk, and retrained no-preview reading score 75.16, 74.38, and 73.91, respectively.

\begin{table}[t]
\caption{HQA SubEM (\%) across context lengths. Avg. and cost use the same five lengths (28K--896K); cost is normalized to Qwen2.5-7B-1M.}
\label{tab:long}
\centering\small
\setlength{\tabcolsep}{4.2pt}
\begin{tabular}{@{}lrrrrrrr@{}}
\toprule
Method & 28K & 112K & 224K & 448K & 896K & Avg.$\uparrow$ & Cost$\downarrow$\\
\midrule
Qwen2.5-7B
& 46.09 & 10.16 & 2.34 & 1.56 & 3.12
& 12.65 & 0.0499$\times$\\

Qwen2.5-7B-1M
& 50.78 & 46.88 & 32.03 & 9.38 & 14.84
& 30.78 & 1.0000$\times$\\

\midrule
BM25-DOS RAG
& 46.09 & 40.62 & 17.97 & 9.38 & 0.78
& 22.97 & 0.0204$\times$\\

Dense-DOS RAG
& 53.91 & 55.47 & 25.78 & 16.41 & 3.91
& 31.10 & 0.0214$\times$\\

BM25 + PGM reader & 79.69 & 64.84 & 70.31 & 75.78 & 72.66 & 72.66 & 0.9960$\times$ \\
Dense + PGM reader & 69.53 & 62.50 & 55.47 & 49.22 & 51.56 & 57.66 & 0.9942$\times$ \\

\midrule
ReadAgent-7B
& 39.06 & 28.91 & 24.22 & 10.16 & 4.69
& 21.41 & 3.1363$\times$\\

MemAgent-7B
& 79.69 & 75.78 & 75.78 & 72.66 & \textbf{75.00}
& 75.78 & 0.8203$\times$\\

LycheeMemory-7B
& 60.16 & 50.00 & 45.31 & 42.19 & 40.62
& 47.66 & 0.8569$\times$\\

ReMemR1-7B
& 82.03 & \textbf{81.25} & 74.22 & 74.22 & 71.88
& 76.72 & 1.1280$\times$\\

InfMem-4B
& 68.75 & 68.75 & 67.97 & 67.19 & 69.53
& 68.44 & 0.3641$\times$\\

MemReread-4B
& 57.03 & 45.31 & 42.97 & 39.84 & 38.28
& 44.69 & 2.4484$\times$\\

\midrule
Raw single-chunk & 78.13 & 77.34 & 72.66 & 76.56 & 71.09 & 75.16 & 0.7252$\times$ \\
Neighboring chunks & 79.69 & 75.00 & 71.88 & 75.78 & 69.53 & 74.38 & 2.9607$\times$ \\
No-preview ReadAll (retrained) & 78.13 & 75.00 & 72.66 & 75.78 & 67.97 & 73.91 & 0.6155$\times$ \\
\midrule
PGM-ReadAll-7B
& 82.03 & 80.47 & 75.78 & 67.97 & 68.75
& 75.00 & 1.8928$\times$\\

\textbf{PGM-Selective-7B}
& \textbf{82.81}
& \textbf{81.25}
& \textbf{76.56}
& \textbf{78.13}
& \textbf{75.00}
& \textbf{78.75}
& 1.0549$\times$\\
\bottomrule
\end{tabular}

\end{table}

\paragraph{LongBench v2.}
\label{sec:lbv2}
In Table~\ref{tab:longbench-v2}, PGM-Selective-7B obtains the highest overall accuracy among the evaluated methods, at 34.79\%, compared with 34.59\% for PGM-ReadAll-7B and 30.42\% for Qwen2.5-7B-1M. Relative to ReadAll, it improves multi-document QA from 27.20\% to 30.40\%, dialogue history from 33.33\% to 41.03\%, and medium-length contexts from 32.56\% to 35.81\%. ReadAll remains stronger on single-document QA and long contexts; both configurations achieve 34.08\% on hard questions. Across this broader task mix, selection preserves comparable overall accuracy with fewer tokens. Relative to MemAgent, the gain is concentrated in the hard subset: accuracy rises from 24.76\% to 34.08\%, while both methods obtain 35.94\% on easy questions.

\begin{table*}[t]
\caption{LongBench v2 accuracy (\%): overall and by task category, context length, and difficulty.}
\label{tab:longbench-v2}
\centering
\small
\setlength{\tabcolsep}{3.1pt}
\resizebox{\textwidth}{!}{
\begin{tabular}{@{}lrrrrrrrrrrrr@{}}
\toprule
&
\multicolumn{1}{c}{Overall} &
\multicolumn{6}{c}{Task category} &
\multicolumn{3}{c}{Context length} &
\multicolumn{2}{c}{Difficulty} \\
\cmidrule(lr){2-2}
\cmidrule(lr){3-8}
\cmidrule(lr){9-11}
\cmidrule(lr){12-13}

Method &
All &
Single-doc &
Multi-doc &
Long-context &
Dialogue &
Code &
Structured &
Short &
Medium &
Long &
Easy &
Hard \\

&
($n$=503) &
QA ($n$=175) &
QA ($n$=125) &
learning ($n$=81) &
history ($n$=39) &
repo. ($n$=50) &
data ($n$=33) &
($n$=180) &
($n$=215) &
($n$=108) &
($n$=192) &
($n$=311) \\
\midrule

Qwen2.5-7B
& 28.83 & 33.71 & 27.20 & \textbf{33.33}
& 12.82 & 32.00 & 12.12
& 36.67 & 25.12 & 23.15
& 32.29 & 26.69 \\

Qwen2.5-7B-1M
& 30.42 & 36.57 & 26.40 & 30.86
& 10.26 & \textbf{40.00} & 21.21
& 39.44 & 24.19 & 27.78
& 35.42 & 27.33 \\

\midrule

BM25-DOS RAG
& 28.23 & 36.57 & 25.60 & 27.16
& 10.26 & 28.00 & 18.18
& 30.56 & 27.91 & 25.00
& 32.81 & 25.40 \\

Dense-DOS RAG
& 30.02 & 36.00 & 28.00 & 25.93
& 20.51 & 32.00 & 24.24
& 35.56 & 26.05 & 28.70
& 31.25 & 29.26 \\

\midrule

ReadAgent-7B
& 29.03 & 36.00 & 28.80 & 25.93
& 25.64 & 20.00 & 18.18
& \textbf{41.11} & 22.33 & 22.22
& 30.73 & 27.97 \\

MemAgent-7B
& 29.03 & 32.00 & 24.80 & 17.28
& 35.90 & 32.00 & 45.45
& 29.44 & 28.37 & 29.63
& \textbf{35.94} & 24.76 \\

LycheeMemory-7B
& 8.75 & 8.57 & 9.60 & 13.58
& 5.13 & 8.00 & 0.00
& 6.67 & 9.77 & 10.19
& 10.94 & 7.40 \\

ReMemR1-7B
& 26.04 & 24.00 & 28.00 & 20.99
& 20.51 & 36.00 & 33.33
& 26.11 & 25.12 & 27.78
& 30.21 & 23.47 \\

InfMem-4B
& 29.22 & 30.29 & 26.40 & 24.69
& 25.64 & 30.00 & \textbf{48.48}
& 29.44 & 26.98 & 33.33
& 32.29 & 27.33 \\

MemReread-4B
& 28.63 & 30.29 & 24.80 & 24.69
& 28.21 & 36.00 & 33.33
& 27.22 & 28.37 & 31.48
& 34.90 & 24.76 \\

\midrule

PGM-ReadAll-7B
& 34.59 & \textbf{38.86} & 27.20 & 32.10
& 33.33 & 36.00 & 45.45
& 35.00 & 32.56 & \textbf{37.96}
& 35.42 & \textbf{34.08} \\

\textbf{PGM-Selective-7B}
& \textbf{34.79}
& 37.71
& \textbf{30.40}
& 29.63
& \textbf{41.03}
& 34.00
& 42.42
& 32.78
& \textbf{35.81}
& 36.11
& \textbf{35.94}
& \textbf{34.08} \\

\bottomrule
\end{tabular}
}
\end{table*}

\subsection{Token efficiency}
\label{sec:quality-cost}

\paragraph{Savings from selective reading.}
Relative to PGM-ReadAll-7B, PGM-Selective-7B reduces total token use by 44.27\% on HQA and 34.76\% on LongBench v2. On HQA, its cost is approximately $0.77\times$ that of ReadAll at 28K and $0.55$--$0.56\times$ at 112K--896K. At 896K, Selective uses 1.65M tokens per query instead of 2.98M, while increasing SubEM from 68.75 to 75.00. The reduced second-pass workload therefore yields substantial end-to-end savings as context length grows.

\paragraph{Comparison with memory-based readers.}
Across the five HQA lengths, Selective improves on ReMemR1 by 2.03 SubEM points while using 6.48\% fewer tokens. ReMemR1 is cheaper at 28K--448K, whereas Selective uses 16.26\% fewer tokens at 896K. Relative to MemAgent, Selective raises average HQA SubEM from 75.78 to 78.75 with 28.61\% more tokens; on LongBench v2, it raises accuracy from 29.03\% to 34.79\% with 26.99\% more tokens. PGM trades additional preview computation for higher answer quality, while learned selection reduces the cost of preview-guided reading.

\paragraph{Raw-text selection and preview reuse.}
PGM uses 64.4\% fewer tokens than neighboring-chunk selection; raw single-chunk selection is cheaper. Reusing previews across ten questions on the same document amortizes their generation cost, reducing per-query tokens by 7.0--26.7\% relative to MemAgent across all five lengths. With four questions, savings are 4.9--14.1\% at 112K--896K.

\subsection{Ablation studies}
\label{sec:ablations}

We study training and selection on HQA at 112K and 224K. Table~\ref{tab:selection-ablation} pairs the answer-quality results with reading ratios and score differences. Random-selection scores are means and standard deviations over five seeds.

\begin{table}[!htb]
\centering
\caption{HQA ablations. Answer scores are SubEM (\%); reading ratios measure second-pass source tokens. Score differences are in percentage points. Panels (c,d) report means over the five HQA context lengths (28K--896K).}
\label{tab:selection-ablation}
\label{tab:readall-ablation}
\label{tab:reading-budget-ablation}
\small
\renewcommand{\arraystretch}{1.13}
\setlength{\tabcolsep}{3pt}
\begin{minipage}[t]{0.49\textwidth}
\centering
\textbf{(a) Answer quality}\par\smallskip
\begin{tabular*}{\linewidth}{@{\extracolsep{\fill}}lrr@{}}
\toprule
Method & 112K & 224K \\
\midrule
\multicolumn{3}{@{}l}{\textit{PGM-Selective}} \\
SFT-only & 78.13 & 74.22 \\
SFT+RL & \textbf{81.25} & \textbf{76.56} \\
Random chunks & $43.91\pm3.64$ & $41.72\pm3.69$ \\
Golden chunks & 87.50 & 85.94 \\
\midrule
\multicolumn{3}{@{}l}{\textit{PGM-ReadAll}} \\
SFT-only & 74.22 & 74.22 \\
SFT+RL & \textbf{80.47} & \textbf{75.78} \\
\bottomrule
\end{tabular*}
\end{minipage}\hfill
\begin{minipage}[t]{0.47\textwidth}
\centering
\textbf{(b) Reading amount and gains}\par\smallskip
\begin{tabular*}{\linewidth}{@{\extracolsep{\fill}}lrr@{}}
\toprule
Metric & 112K & 224K \\
\midrule
\multicolumn{3}{@{}l}{\textit{Reading ratio (\%)}} \\
SFT+RL & 46.22 & 44.16 \\
Random & 45.66 & 43.89 \\
\midrule
\multicolumn{3}{@{}l}{\textit{SubEM difference (pp)}} \\
RL gain (Selective) & $+3.13$ & $+2.34$ \\
RL gain (ReadAll) & $+6.25$ & $+1.56$ \\
Learned $-$ Random & $+37.34$ & $+34.84$ \\
Golden $-$ Learned & $+6.25$ & $+9.38$ \\
\bottomrule
\end{tabular*}
\end{minipage}
\par\medskip
\begin{minipage}[t]{0.49\textwidth}
\centering
\textbf{(c) Preview conditioning}\par\smallskip
\begin{tabular*}{\linewidth}{@{\extracolsep{\fill}}lrr@{}}
\toprule
Selector & With preview & Without preview \\
\midrule
PGM & \textbf{78.75} & 76.09 \\
BM25 & 72.66 & 69.84 \\
Dense & 57.66 & 55.78 \\
\bottomrule
\end{tabular*}
\end{minipage}\hfill
\begin{minipage}[t]{0.47\textwidth}
\centering
\textbf{(d) Final-memory retention}\par\smallskip
\begin{tabular*}{\linewidth}{@{\extracolsep{\fill}}lr@{}}
\toprule
Method & Gold facts retained (\%) \\
\midrule
MemAgent & 65.51 \\
PGM-ReadAll & 86.25 \\
PGM-Selective & 85.57 \\
\bottomrule
\end{tabular*}
\end{minipage}
\label{tab:preview-conditioning}
\label{tab:retention-summary}
\end{table}

\paragraph{Learning from answer feedback.}
With the reader fixed, RL improves the selector's SubEM from 78.13 to 81.25 at 112K and from 74.22 to 76.56 at 224K, gains of 3.13 and 2.34 percentage points. Final-answer feedback complements supporting-fact supervision by evaluating whether the selected chunks support a correct answer after memory updates.

\paragraph{Selecting useful content.}
SFT+RL reads 46.22\% of source tokens at 112K and 44.16\% at 224K, close to the random baseline's 45.66\% and 43.89\%. Yet its SubEM is higher by 37.34 and 34.84 points. Which chunks are read, rather than reading less alone, is central to the result. Selecting golden chunks yields 87.50 and 85.94 SubEM, leaving gaps of 6.25 and 9.38 points to learned selection.

\paragraph{Training the full reader.}
The PGM-ReadAll block in Table~\ref{tab:selection-ablation} compares its SFT-only and SFT+RL results. RL improves SubEM by 6.25 points at 112K and 1.56 points at 224K. Together, the two training comparisons show gains from answer feedback for both full reading and learned selection.

\paragraph{Preview conditioning.}
With the reader and selected chunks fixed, preview conditioning improves PGM's reported average SubEM by 2.66 points (Table~\ref{tab:preview-conditioning}). The separately retrained no-preview ReadAll baseline averages 73.91, compared with 78.75 for PGM-Selective.

\begin{figure}[!htb]
\centering
\includegraphics[width=0.95\linewidth]{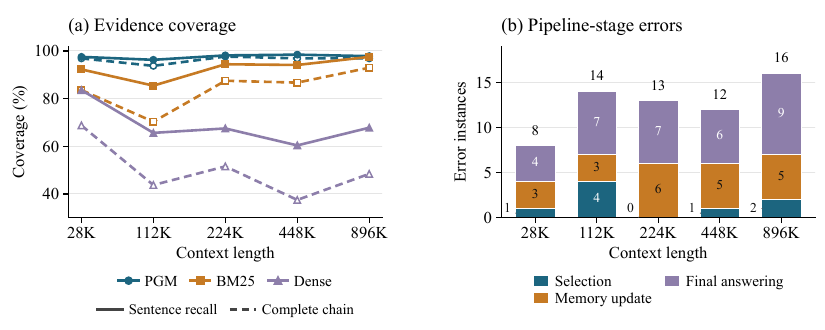}
\caption{\textbf{Evidence coverage and error attribution.} Left: supporting-sentence recall (323 gold sentences) and complete-chain coverage (128 questions) at approximately matched reading budgets. A complete chain includes all supporting sentences and source titles. Right: 63 pipeline-stage errors within an audit of 135 incorrect instances across five lengths.}
\label{fig:coverage-errors}
\end{figure}

\paragraph{Evidence preservation.}
PGM selects 96.28--98.45\% of supporting sentences and complete chains for 93.75--97.66\% of questions (Figure~\ref{fig:coverage-errors}). At 896K, these remain 97.83\% and 96.88\%. The audit attributes 8 failures to selection, 22 to memory updates, and 33 to final answering. Selective retains comparable gold facts to ReadAll (Table~\ref{tab:retention-summary}d) while reducing mean memory length by 9.29\%, the irrelevant-token fraction by 3.24 points, and update calls by approximately 55\% at 112K--896K. Detailed memory statistics are provided in Table~\ref{tab:memory-diagnostics} in Appendix~\ref{app:memory-diagnostics}.


\section{Conclusion}
\label{sec:conclusion}

PGM learns which source chunks a bounded-memory reader should consolidate. Indexed previews support admission and memory updating, while terminal answer feedback evaluates the selected source set through a fixed reader. Relative to PGM-ReadAll, PGM-Selective improves HQA answer quality and reduces token use, while maintaining comparable accuracy at lower token cost on LongBench v2. Learning source admission is therefore an effective approach to allocating inference tokens for long-context reasoning. Incremental preview generation for dynamically growing contexts is a direction for future work.

\label{main-text-end}
\clearpage
\section*{AI use statement}
We used an AI assistant to support manuscript preparation and improve the clarity of the writing. The authors take full responsibility for the content and conclusions.

\begingroup
\setlength{\bibsep}{6pt plus 1pt minus 1pt}
\bibliography{references}
\bibliographystyle{iclr2027_conference}
\endgroup
\clearpage
\appendix
\raggedbottom
\section{Implementation details}
\label{app:implementation}
\label{app:prompts}

\paragraph{Prompt interfaces.}
Templates are included in \texttt{anc/prompts/}. The preview prompt requests ordered topics without the question. The selector receives the question, page outline, focused entries, and decision slots; the update prompt receives the raw chunk, question, memory, local section index, and page outline. The final prompt uses only the question and memory and requests a boxed answer. Pages split recursively to fit the input budget, preserving complete entries.

\paragraph{Decision slots.}
At each \texttt{<gate\_slot>} end marker, fixed vocabulary output rows $w_R,w_S$ give the logit $\ell_i=(w_R-w_S)^\top h_i+b_\phi$ and score $a_i=\sigma(\ell_i)$. Scores are extracted directly in one forward pass per page. No autoregressive decision tokens are generated.

\paragraph{Inference selection.}
Given threshold $\tau$, let $S_\tau=\{i:a_i\geq\tau\}$ and let $\mathcal F$ contain chunks whose previews reach the output cap. Inference includes these chunks and falls back to the highest-scoring chunk if the resulting set is empty:
\begin{equation}
 \widetilde S=S_\tau\cup\mathcal F,\qquad
 S=\begin{cases}
 \widetilde S,&\widetilde S\ne\varnothing,\\
 \{\arg\max_i a_i\},&\widetilde S=\varnothing.
 \end{cases}
 \label{eq:selection-safeguards}
\end{equation}
The highest-score fallback applies at inference. Training can sample an empty set when no chunk is forced.

\paragraph{Memory-update runtime.}
In Equation~\ref{eq:reader}, a standalone \texttt{<KEEP\_MEMORY>} response preserves memory after whitespace stripping and case normalization. Other responses replace it, including those reaching the output cap. Memory carries across preview pages.

\paragraph{Retrieval controls.}
BM25 ($k_1=1.2$, $b=0.75$) and Qwen3-Embedding-0.6B with cosine similarity rank 5,000-token raw chunks. We choose the nonempty ranked prefix plus shared forced chunks closest to each question's PGM raw-token budget, then restore source order. Both use the frozen PGM reader (Stage~II, step~5), identical preview pages, and a 1,024-token memory limit.

\section{Training settings}
\label{app:training}

\paragraph{Trainable parameters.}
During selective SFT and selective RL, we update the selector's attention LoRA adapters~\citep{hu2022lora} and scalar bias while keeping the reader fixed. Objectives are defined in Section~\ref{sec:training}.

\paragraph{Evidence alignment.}
For supporting fact $e_j$, let $\mathcal A_j$ contain its alternative source occurrences. Each occurrence is a set of chunk indices containing the full fact, including every chunk needed by a boundary-crossing sentence. Alignment uses question/answer identity, document title, and sentence offsets. The labels in Equation~\ref{eq:bce} and coverage indicators in Equation~\ref{eq:coverage} are
\begin{equation}
 y_i^{\mathrm{ev}}=\mathbf1\!\left[i\in\bigcup_j\bigcup_{A\in\mathcal A_j}A\right],
 \qquad g_j(S)=\mathbf1[\exists A\in\mathcal A_j:A\subseteq S].
 \label{eq:evidence-alignment}
\end{equation}

\paragraph{Objective coefficients.}
We use $w_+=3$ in Equation~\ref{eq:bce}; $(\alpha,\beta,\gamma,\delta,\eta,\zeta)=(2,2,0.1,0.2,0.5,0.25)$ in Equation~\ref{eq:reward}; and $(\lambda_{\mathrm{KL}},\lambda_{\mathrm{ev}},\lambda_{\mathrm{ent}})=(0.05,0.02,0.002)$ in Equation~\ref{eq:loss}.

\paragraph{Sampling and reading bonus.}
For each nonforced chunk, $p_i=0.04+0.92a_i$. Chunks in $\mathcal F$ from Appendix~\ref{app:implementation} are always selected, so $\mathcal I=\{1,\ldots,N\}\setminus\mathcal F$ in Equation~\ref{eq:set-probability}. The reading-bonus schedule in Equation~\ref{eq:reward} is $\lambda_u=0.25\min(1,u/32)$ for training updates $u=1,2,\ldots$. All component scores use a 0--1 scale.

\paragraph{Loss aggregation.}
In Equation~\ref{eq:loss}, the KL term compares the transformed Bernoulli probabilities with those of the supervised reference. KL, entropy, and evidence regularization include forced chunks and average over chunks. Advantages are held fixed when differentiating the policy loss. The final loss averages over questions, with one selector update per sampled batch.

\clearpage
\section{Paired trajectory case study}
\label{app:case-study}

\paragraph{Question and runs.}
The motivating example in Figure~\ref{fig:hqa-case} is HQA source \texttt{5ae63dad55429929b0807afe} at 112K tokens. It asks: \emph{When was the British author who wrote the novel on which ``Here We Go Round the Mulberry Bush'' was based born?} The correct answer is 7 January 1936. The paired runs use the same source document; the three key chunks (11, 16, and 22) were checked to be identical despite differences in the full-document chunking implementations. The MemAgent trajectory uses the public MemAgent-7B weights and original single-pass prompts in an independent diagnostic run. The PGM trajectory uses the supervised selector before reinforcement learning (selector step~2944 and reader \texttt{stage2/global\_step\_5}). The MemAgent trajectory follows the sampled evaluation protocol, whereas PGM uses greedy decoding. These case-study runs do not replace the benchmark results.

\paragraph{Visible evidence and memory.}
Chunk~11 states that Edward Hunter Davies was born on 7 January 1936. Chunk~16 contains the same-title nursery rhyme, and chunk~22 identifies the film as an adaptation of Davies's novel. PGM selects chunks 4, 8, 9, 11, 13, 16, and 22, covering 31.38\% of raw-source tokens in the second pass. Between chunks~11 and~22, MemAgent performs ten intervening updates; PGM reads only chunks~13 and~16. Its page at chunk~11 covers chunks 1--12; the page at chunk~16 covers 13--23 and lists the later film title without the author relation or birth date. PGM retains the date at chunk~16 and adds the author relation at chunk~22. Both facts occur in its final memory. Its answer is 7 January 1936; the original MemAgent answer is February 24, 1936. The diagnostic records preserve the source passages, actual preview pages, selection decisions, and intermediate memories.

\paragraph{Local rewrite control.}
Replaying PGM reproduces all seven memory updates and the final answer. We then require an explicit rewrite at chunk~16, fixing all other inputs, selected chunks, reader weights, preceding updates, budgets, and greedy decoding. An instruction-only attempt still returns \texttt{<KEEP\_MEMORY>}. The effective intervention combines the rewrite instruction with a ban on the first-token \texttt{<} entry used by KEEP; it also excludes other outputs starting with that token, without constraining subsequent tokens or inserting evidence. Chunk~22 and final answering use their original prompts with the resulting memory. The model rewrites the memory, retains the birthday, and answers correctly. Evidence retention in this case therefore does not depend on taking the explicit KEEP branch.

\paragraph{Delayed geographical association.}
In HQA 112K question~50 (zero-indexed), MemAgent drops Sullivan County's seat, Newport, after chunk~17 and answers North Haverhill. PGM retains Newport until chunk~22 links East Lempster to Sullivan County and answers correctly. Its preview lists later Lempster entries without Newport. These illustrative trajectories use their native readers and decoding protocols.

\section{Final-memory diagnostics}
\label{app:memory-diagnostics}
\begin{table}[!htb]
\centering
\caption{Final-memory diagnostics over 128 questions per length. Retention is the fraction of 323 gold facts fully preserved (\%). Paired entries are ReadAll / Selective, with a shared 1,024-token cap. Lengths are in tokens; overall length statistics pool all 640 instances. Calls include KEEP responses.}
\label{tab:memory-diagnostics}
\fontsize{8}{9}\selectfont
\setlength{\tabcolsep}{2.1pt}
\renewcommand{\arraystretch}{1.03}
\begin{tabular}{@{}lrrrrrr@{}}
\toprule
Metric / method & 28K & 112K & 224K & 448K & 896K & Overall \\
\midrule
Retention: MemAgent & 67.49 & 70.59 & 66.87 & 62.54 & 60.06 & 65.51 \\
Retention: ReadAll & 82.66 & 87.93 & 86.69 & 87.31 & 86.69 & 86.25 \\
Retention: Selective & 80.80 & 86.38 & 86.38 & 87.62 & 86.69 & 85.57 \\
\midrule
Mean length & 151.56/143.26 & 276.32/239.69 & 317.39/284.16 & 373.59/331.03 & 434.42/410.84 & 310.66/281.80 \\
Median length & 138.0/129.5 & 238.0/210.0 & 296.0/254.0 & 348.5/315.5 & 380.5/370.5 & 260.0/234.0 \\
P5 length & 54.40/54.40 & 87.90/81.35 & 106.10/83.40 & 139.70/112.45 & 148.75/135.80 & 79.95/72.95 \\
P95 length & 278.70/278.70 & 563.50/488.50 & 682.10/625.60 & 711.40/589.10 & 875.30/829.20 & 701.00/645.50 \\
Irrelevant tokens (\%) & 29.90/28.08 & 50.20/45.72 & 55.59/50.23 & 61.60/58.47 & 65.84/64.43 & 52.63/49.39 \\
Update calls / query & 6.05/3.55 & 22.95/10.40 & 45.50/19.87 & 90.79/40.05 & 181.13/80.43 & 69.28/30.86 \\
\bottomrule
\end{tabular}
\end{table}

\end{document}